\documentclass{article}

\usepackage{arxiv}
\usepackage{balance} 
\usepackage{caption}
\usepackage{subcaption}
\usepackage{mathtools}
\usepackage{algorithm}
\usepackage{algpseudocode}
\usepackage[utf8]{inputenc} 
\usepackage[T1]{fontenc}    
\usepackage{hyperref}       
\usepackage{url}            
\usepackage{booktabs}       
\usepackage{amsfonts}       
\usepackage{nicefrac}       
\usepackage{microtype}      
\usepackage{lipsum}		
\usepackage{graphicx}
\usepackage{natbib}
\usepackage{doi}
\usepackage{multirow}
\usepackage{amsthm}
\usepackage{array}
\newtheorem{definition}{Definition}

\title{Using Non-Stationary Bandits for Learning in Repeated Cournot Games with Non-Stationary Demand}

\author{ Kshitija Taywade \\
	Department of Computer Science\\
	University of Kentucky\\

	\And
    Brent Harrison \\
	Department of Computer Science\\
	University of Kentucky\\
	
	\And
    Judy Goldsmith \\
	Department of Computer Science\\
	University of Kentucky}

\hypersetup{
pdftitle={Using Non-Stationary Bandits for Learning in Repeated Cournot Games with Non-Stationary Demand},
pdfsubject={Multi-armed Bandits},
pdfauthor={Kshitija Taywade, Brent Harrison, Judy Goldsmith},
pdfkeywords={Multi-agent Learning, Multi-armed Bandits, Non-stationary Bandits, Cournot Games},
}

\begin{document}

\maketitle 

\begin{abstract}
Many past attempts at modeling repeated Cournot games assume that demand is stationary. This does not align with real-world scenarios in which market demands can evolve over a product's lifetime for a myriad of reasons. In this paper, we model repeated Cournot games with non-stationary demand such that firms/agents face separate instances of non-stationary multi-armed bandit problem. The set of arms/actions that an agent can choose from represents discrete production quantities; here, the action space is ordered. Agents are independent and autonomous, and cannot observe anything from the environment; they can only see their own rewards after taking an action, and only work towards maximizing these rewards. We propose a novel algorithm \textit{Adaptive with Weighted Exploration (AWE) $\epsilon$-greedy} which is remotely based on the well-known $\epsilon$-greedy approach. This algorithm detects and quantifies changes in rewards due to varying market demand and varies learning rate and exploration rate in proportion to the degree of changes in demand, thus enabling agents to better identify new optimal actions. For efficient exploration, it also deploys a mechanism for weighing actions that takes advantage of the ordered action space. We use simulations to study the emergence of various equilibria in the market. In addition, we study the scalability of our approach in terms number of total agents in the system and the size of action space. We consider both symmetric and asymmetric firms in our models. We found that using our proposed method, agents are able to swiftly change their course of action according to the changes in demand, and they also engage in collusive behavior in many simulations.

\end{abstract}

\keywords{Multi-armed Bandits, Non-stationary Bandits, Cournot Games}

\section{Introduction}

In a standard Cournot game \citep{cournot1838recherches}, firms compete over the production of identical goods. Various real-world markets can be  modeled as Cournot games; for example, energy systems \citep{kirschen2018fundamentals}, transportation networks \citep{bimpikis2019cournot}, and healthcare systems \citep{chletsos2019hospitals}. Another closely related problem is dynamic pricing. Most of the literature on Cournot games assumes stationary demand functions. This, however, does not align with many real-world market settings as they are mostly non-stationary due to factors such as changing seasons, trends, and as seen most recently, global health crises. For example, the decision problem of fresh food productions where products have limited shelf life after which the products cannot be sold; the demand for such products is always fluctuating due to various reasons, one of them being promotional offers by retailers.
There has been several studies on  non-stationary demand in such inventory management models \citep{graves2008strategic,silver2008inventory,tunc2011cost,yue2010selecting,mohebbi2005impact}.

Along with environmental factors, activities of other competing firms can also affect market demand from a single firm's perspective. This assumption that demand is stationary, therefore, limits the general applicability of past approaches for solving Cournot games in real-world settings. To address this limitation of past approaches, we consider repeated Cournot games with non-stationary demand functions. Decision making in these environments can be very difficult for an autonomous and independent firm. In real-world markets, firms often do not have access to knowledge about market demands or their competitors' behavior. This also contradicts many of the assumptions made by prior works. To better model these situations, we consider Cournot games such that firms (agents) are not able to directly observe information about other agents in the environment or the demand function. The only information available to an individual agent is the profit obtained after selecting a production quantity, which is a single play of a Cournot game where firms produce simultaneously. We consider three different types of non-stationarities in the demand. Demand patterns are inspired from the works based on economic markets \citep{berry1972lot,tunc2011cost}. We cover both sudden and incremental changes, as well as reoccurring and uncertain changes. Firms are not able to access any information on the market demand.

In literature, Cournot games have been modeled in various ways with different assumptions on the firms' cognitive capabilities and rationalities. Several works analyze characteristics of learning mechanisms required to reach a specific equilibrium. Many learning mechanisms assume that the firms have some knowledge of the market and their competitors. The problem of managing production while facing changing demand becomes harder when coupled with a lack of any information. In this paper, we model repeated Cournot games where firms learn independently and are autonomous. To assist individual agents in learning and making decisions, we model the problem as non-stationary bandits. Each agent deals with its own non-stationary multi-armed bandit problem separately. In any multi-armed bandits (MAB) problem, dealing with the exploration-exploitation trade-off is a crucial task. In our multi-agent setting, agents learn simultaneously. As learning involves exploration, high uncertainty in the rewards can be seen from the perspective of a single agent having no knowledge of the environment, regardless of the demand being stationary or otherwise. We consider market settings where random entry or exit is not allowed to firms. Moreover, we assume that the agents use the same learning mechanism. Due to these assumptions, we consider the problem as non-stationary bandits rather than adversarial bandits.

We propose a novel approach called \textit{Adaptive with Weighted Exploration (AWE) $\epsilon$-Greedy}. It is remotely based on the well-known $\epsilon$-greedy strategy, as $\epsilon$-greedy can work with our assumptions of the lack of knowledge, while some other popular methods (for example, UCB and Thompson sampling) may require some additional knowledge from the environment. This approach makes use of the fact that the action space is ordered in Cournot models. It incorporates the mechanism to quantify changes in rewards. According to the changes in reward, it make changes in exploration rate as well as learning rate; this helps in finding the new optimal arm. Additionally, it tries to optimize the exploration by assigning weights to the arms and choosing the arms as per those weights; these weights are dependant on the current optimal arm and the degree of observed changes quantified using another simple technique. These techniques also help in dealing with the uncertainty in rewards caused by the activities of other agents operating in the system.

We evaluate our proposed approach empirically by running different kinds of simulations. We investigate the performance of our approach in terms of responsiveness to changing demand. We use adaptive $\epsilon$-greedy algorithm \citep{dos2017adaptive} as a baseline to compare our results; this algorithm is somewhat similar to our approach as it is also based on the $\epsilon$-greedy strategy. We check our outcomes in relation to three main types of equilibria that are associated with Cournot games: Walrasian equilibrium, Cournot/Nash equilibrium, and collusive equilibrium. We investigate the scalability in terms of the number of firms in the system and the size of the action space. We also extend our simulations to Cournot models with asymmetric firms, i.e., firms with different marginal costs. We start with simple small-scale simulations same as that used in \citep{waltman2008q,xu2020reinforcement}; the only difference is that we consider non-stationary demand functions. We found that by using our approach, agents are able to readily change their action course according to changing demand. Agents show collusive behavior, although full collusion usually does not emerge. By collusive behavior, we mean outputs anywhere between Nash and collusive equilibrium. With an increasing number of firms in the system, the collusive behavior declines. However, the results still indicate the possibility of the emergence of collusive behavior, even without explicit instructions to collude. We have also added a brief analysis of our results with the notion of joint cumulative regret. In models with symmetric firms, individual agents produce similar quantities; however, asymmetry in firms can cause unfair outcomes with substantial differences in the performance of different agents.


\section{Related Work}

Literature based on learning methods in Cournot games mostly assumes stationary demand. However, in real-world scenarios, non-stationarity is integral to economic markets. Depending on the type of market, changes in demand follows various patterns, i.e., small and slow changes to large and sharp changes. Several works study the inventory management model where companies have to decide efficiently about the production or stocking of goods \citep{graves2008strategic,silver2008inventory,tunc2011cost,yue2010selecting,mohebbi2005impact}. 

For Cournot games, the learning of economic agents can be widely divided into two categories: individual learning and social learning \citep{vriend2000illustration}. The convergence to various equilibria have also been widely studied in past research. The literature establishes that the long-term outcome of a Cournot oligopoly model depends on the underlying learning mechanism, firms' rationality, and the memory size of the firms. In individual learning models, an agent learns exclusively from its own experience, whereas in social learning models, an agent also learns from the experience of other agents. Learning frameworks also reflect the rationality of agents. Several works have studied individual learning mechanisms \citep{vriend2000illustration,arifovic2006revisiting,vallee2009convergence,fudenberg1998theory,riechmann2006cournot}. Some works claim that while social learning leads to Walrasian equilibrium, individual learning scheme results in convergence of the Nash equilibrium \citep{vriend2000illustration,vega1997evolution,franke1998coevolution,dawid2011adaptive,bischi2015evolutionary}. In this work, we model the problem of decision making by a single agent in a repeated Cournot game with non-stationary demand as a non-stationary multi-armed bandit problem. Firms make decisions based on the profit/reward they get, mostly unaware of the impact of other firms' actions on the market price; this can be called as implicit individual learning. While learning in MAB setting, no assumptions on knowledge of the environment are required. Hence, we think that this is the better way to model real-world scenarios.

Non-stationary multi-armed bandits are mainly divided into two categories: rested bandits \citep{gittins1979dynamic,bouneffouf2014contextual,bouneffouf2016multi,levine2017rotting,seznec2020single} and restless bandits \citep{gafni2018learning,liu2012learning,meshram2018whittle,besson2018doubling,cheung2019learning,russac2019weighted,wei2016tracking,seznec2020single}. In rested bandit case, the underlying distribution changes only when the arm is played. While in the case of restless bandits, the underlying distribution of all the arms changes at every time step according to known but arbitrary stochastic transition function. Additionally, some problems are categorized as piece-wise stationary \citep{garivier2011upper,yu2009piecewise,cao2019nearly} in which the reward distributions for the arms are piece-wise stationary and will shift at some unknown time steps. Several approaches have been proposed to deal with different kinds of non-stationarities. Some approaches are based on UCB algorithm \citep{garivier2011upper} and some others on Thompson sampling method \citep{ghatak2020change}. Additionally, there are methods based on change-point detection; for example, Change-Point Thompson Sampling \citep{mellor2013thompson} and Change Detection UCB \citep{liu2018change}. An approach of forgetting past memory to deal with non-stationarity has also been proposed \citep{russac2019weighted}. Many of these approaches are designed for problems different than ours; they also make different assumptions on models.

Our approach is remotely based on $\epsilon$-greedy algorithm. Closely related to our work is the work by \citeauthor{dos2017adaptive} in which authors propose adaptive $\epsilon$-greedy algorithm which modifies exploration rate based on changes detected in rewards \citep{dos2017adaptive}. Our algorithm employs different mechanism to quantify changes than the ones used in this work; in addition, our algorithm uses weighted exploration technique; this technique is specific to Cournot games where action set is ordered.

Researchers have also used other RL methods to facilitate learning in repeated Cournot games \citep{kimbrough2003note,waltman2008q,xu2020reinforcement}, but as per our knowledge, all of them consider stationary demand. \citeauthor{waltman2008q} analyze the results of Q-learning in a Cournot game with discrete action space and explain the emergence of collusive behavior. They focus on three types of firms; firms in our experiments are similar to the firms without memory as considered in their paper. \citep{kimbrough2003note} also reported results on Q-learning behavior in a Cournot oligopoly game where they found a slight tendency towards collusive behavior in their simulation study. \citeauthor{xu2020reinforcement} incorporated memory and imitation with RL. In their model, firms do not have any information about market demand, but they can observe quantities produced by the other firms and their profits. They use a continuous action space and parametric function approximation. They use three different settings, the first of which (Treatment 1 in their paper) resembles our models. They have used the same experimental setup as in \citep{waltman2008q}. Unlike our results, they observe convergence to Nash equilibrium for these settings. Both of these papers evaluate scalability up to a limited range, while our paper incorporates a more thorough investigation of scalability in terms of the number of firms and the size of action space. \citeauthor{huck2004through} studied another trial-and-error method. In their work, firms do not have information about their rivals as well as the payoff function of the game. However, their method allows agents only to decrease or increase the production level; therefore, there is a great deal of responsibility on firms to choose the starting quantity. They showed that full collusion can be achieved with their approach; however, they did not study the scalability of their approach. There is a wide variety of work that uses RL and MAB approaches for dynamic pricing problems; the market models considered in these works are diverse \citep{den2015dynamic,kephart2000pseudo,kononen2006dynamic,misra2019dynamic,hansen2020algorithmic,trovo2015multi}. 


\section{Preliminaries}

\subsection{Cournot Games}

We consider a standard Cournot model with $n$ firms offering an identical product. Firms produce independently and simultaneously and compete on the quantity they produce. In our model, the product quantity is a discrete number. Firm $i$ produces quantity $q_{i}$ at each time step
$t$. Firm $i$'s total cost is $C_{i} = cq_{i}$, where $c$ is the constant marginal cost. The linear demand function with an inverse demand equation is

\begin{equation}\label{eq1}
    p = max(u - v\sum_{i=1}^{n}q_{i},0)
\end{equation}
where $u>0$ and $v>0$ denote two parameters. This is the stationary demand function when parameters $u$ and $v$ are constant. In our models, we still use the same function to calculate the demand, however, parameter $u$ is no more a constant value. We change it throughout the game to add non-stationarity in the market demand.

The profit of a firm $i$ is calculated as

\begin{equation}\label{eq2}
\pi_{i} = p q_{i} - C_{i} \ for \ i=1,\dots,n.
\end{equation}

\subsection{Non-stationarity in Demand}

\begin{figure*}
     \centering
     \begin{subfigure}[b]{0.3\textwidth}
         \centering
         \includegraphics[width=\textwidth]{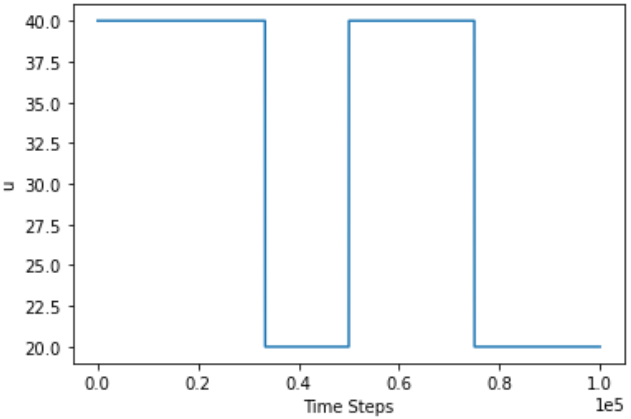}
         \caption{}
         \label{fig:dp1}
     \end{subfigure}
     \hfill
     \begin{subfigure}[b]{0.3\textwidth}
         \centering
         \includegraphics[width=\textwidth]{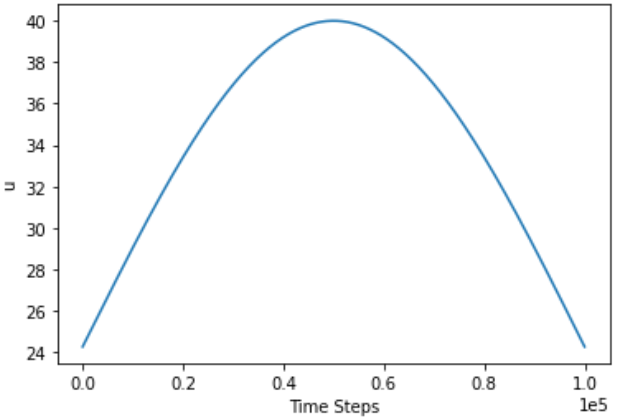}
         \caption{}
         \label{fig:dp2}
     \end{subfigure}
     \hfill
     \begin{subfigure}[b]{0.3\textwidth}
         \centering
         \includegraphics[width=\textwidth]{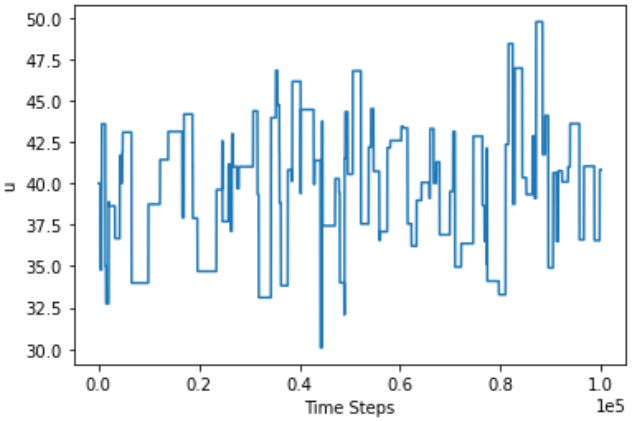}
         \caption{}
         \label{fig:dp3}
     \end{subfigure}
        \caption{Examples of three different demand patterns. X-axis represents time steps in the simulation and Y-axis represents varying values of parameter $u$ (from the inverse demand function $u-vQ$, presented in eq. \ref{eq1}). Here, default value for $u$, $u_{s}=40$. Figs. \ref{fig:dp1}, \ref{fig:dp2}, and \ref{fig:dp3} shows demand patterns 1, 2, and 3, resp.}
        \label{fig:dp}
\end{figure*}

We add three different types of non-stationarities in the underlying demand presented in eq. \ref{eq1}. As mentioned above, we make changes in parameter $u$ to make the demand non-stationary. Parameter $v$ is always constant ($v=1$) in our models, therefore $u$ dictates the changes in demand. The patterns emerging from these changes are shown in fig. \ref{fig:dp}. These patterns are inspired from the works based on economic markets \citep{berry1972lot,tunc2011cost}. We have chosen to explore these patterns as they incorporate some of the common non-stationary behaviors of demand seen in real world industrial settings, i.e., sudden changes (shocks), slow changes, and more erratic changes in demand. We refer to these demand patterns as patterns 1, 2, 3, resp. Unlike other works, we derive the demand pattern solely by changing $u$. At the start of the game, $u$ is initialized with some value, $u_{s}$, and the changes are made to this initial value to change the demand. Fig. \ref{fig:dp1} represents sudden and reoccurring changes in market demand; $u_{s}$ changes according to following rules:
\begin{center}
\[ 
  u_{t} = 
  \begin{dcases*} 
  \text{$u_{s}$/2,} & when t=T/3 \\ 
  \text{$u_{s}$,} &  when t=T/2 \\ 
  \text{$u_{s}$/2,} & when t=3T/4 \\
  \end{dcases*} 
\]
\end{center}
where $T$ is total number of time steps in a simulation of repeated Cournot game. For the time steps other than the ones mentioned above, $u_{t}$ remain stationary, i.e., $u_{t}=u_{t-1}$. Fig. \ref{fig:dp2} represents sinusoidal pattern with demand changing continuously; there are no sudden changes, instead the demand changes incrementally and then gradually decreases with small variations. $u_{t}$ is calculated as follows:

\begin{center}
$u_{t}=\frac{u_{s}}{f(T/2|T/2, T/2)} \cdot f(t | T/2, T/2)$
\end{center}
where $f$ represents normal probability density function. Both the demand patterns discussed above are dynamic but deterministic. Unlike these patterns, the third pattern, presented in fig. \ref{fig:dp3} is stochastic. It is an example of erratic demand pattern which is extreme case of non-stationarity. Here, $u_{t}$ is calculated as follows:

\begin{center}
\[ 
  u_{t} = 
  \begin{dcases*} 
  \text{$u_{t-1} \cdot |X|$,} & if $z_{t}<\gamma$ \\ 
  \text{$u_{t-1}$,} &  otherwise \\ 
  \end{dcases*} 
\]
\end{center}
where $X \sim \mathcal{N}(\mu,\,\sigma^{2})$, with $\mu=1$ and $\sigma=0.2$; $z_{t} \sim \mathcal{U}(0,\,1)$. $\gamma=0.01$; it dictates the probability of change in demand at any time step $t$.

\subsection{Non-stationary Multi-armed Bandit Framework}

We consider $n$ firms in repeated Cournot game. Let $K_{i}$ be the set of arms/actions faced by agent $i$; arms represent the different production choices. The action space is discrete and ordered; every action represents a production level choice, and the action set consists of integer values in a specific range. Agents do not have any knowledge of the environment; agents only have knowledge about their own payoffs/rewards after taking an action. As agents are independent and autonomous, they each face a separate non-stationary MAB problem. Each agent's goal is to maximize their individual reward/profit over a long (possibly infinite) time horizon $T$. At time step $t$, the reward is calculated using eq. \ref{eq2}. From eq. \ref{eq1} and \ref{eq2}, we can see that reward not only depends on demand (or parameter $u$), but also depends on individual agent's production as well as total industry output at time step $t$. This adds more uncertainty to underlying non-stationary nature of rewards.

\subsection{Equilibria in Cournot Oligopoly Models}

There are three main equilibria associated with Cournot games. The following definitions are for models with symmetric firms.

\begin{definition}
The \textbf{Cournot (Nash) Equilibrium} is obtained if each firm chooses the production level that maximizes its profit, given the production levels of its competitors. No firm wishes to unilaterally change its output level when the other firms produce the output levels assigned to them in the equilibrium. Firms individually maximize their profit; they do not maximize their joint profit. The resulting joint production level is $\frac{(u-c)n}{v(n+1).}$

\end{definition}

\begin{definition}
\textbf{Walrasian Equilibrium} is obtained if firms are not aware of their influence on the market price, and therefore behave as price takers. They adopt Walrasian rule and produce Walrasian quantity. The Walrasian rule is based on the assumption that a firm acting as price taker decides next-period output maximizing its profit. The resulting quantity dynamics leads to a dynamic equation that allows the Walrasian equilibrium output as the unique steady state \citep{radi2017walrasian}. The resulting joint production level is $\frac{(u-c)}{v}.$

\end{definition}

\begin{definition}
In \textbf{Collusive Equilibrium}, firms form a cartel and maximize the joint profit. They produce a smaller quantity than the quantity that maximizes their individual profit. Hence, they have an incentive to increase their production levels. The resulting joint production level is $\frac{(u-c)}{2v}.$

\end{definition}

\section{Method}

We have used the non-stationary bandit problem to model the decision making in repeated Cournot games with non-stationary demand. Each firm in the market separately faces its own instance of non-stationary MAB problem. Rewards associated with arms are based on market demand as well as actions taken by other agents in the system. Non-stationary demand means that the rewards are also inherently non-stationary. As each agent is exploring and learning, the actions selected by other agents in the environment are not constant, which adds further uncertainty in the rewards obtained by an individual agent which cannot see anything from the environment, except its own rewards.

We propose a novel approach to deal with non-stationarity in rewards specifically designed for environments like Cournot models where the action space is ordered. The proposed approach takes advantage of this feature of the action space. It also deals with the uncertainty in rewards as a result of multiple agents operating in the system. This method is remotely based on the well-known $\epsilon$-greedy approach, as $\epsilon$-greedy approach can work with our assumptions of the lack of knowledge, while some other popular methods (for example, UCB and Thompson sampling) may require some additional knowledge from the environment. Even though conventional $\epsilon$-greedy is not considered efficient and applicable to non-stationary environments, our proposed novel exploration mechanism along with change detection mechanism is designed to address its drawbacks. Subsequently, we get simplistic method like $\epsilon$-greedy but with capability of efficiently handling non-stationarity.

\subsection{$\epsilon$-Greedy \citep{sutton1998reinforcement}}
$\epsilon$-Greedy is a common method for balancing exploration and exploitation trade-offs. At each time step $t=1,2,..$, an agent chooses a random action with probability $\epsilon$ or otherwise chooses the action with the highest empirical mean. The empirical mean of rewards after taking an action $a$ is often referred to as that action's $Q$-value and is denoted as $Q_{t}(a)$ for time step $t$. However, in our formulation, $Q_{t}(a)$ is calculated using learning rate $\alpha$ (inspired by Q-learning algorithm \citep{sutton1998reinforcement}). It is calculated as follows:
\begin{center}

    $Q_{t}(a) = \alpha R_{t} + (1-\alpha) Q_{t-1}(a)$

\end{center}

where $R_{t}$ is the reward obtained at time step $t$. A linear bound on the expected regret can be achieved with constant $\epsilon$. For variant of the algorithm where $\epsilon$ decreases with time, \citep{cesa1998finite} proved poly-logarithmic bounds. However, \citep{vermorel2005multi} did not find any practical advantage to using these methods in their empirical study. Here, we are referring to regret as it is used conventionally in single agent stochastic MAB problems \citep{kuleshov2014algorithms,lattimore2020bandit}.

\subsection{Adaptive with Weighted Exploration (AWE) $\epsilon$-Greedy}

This approach is based on $\epsilon$-greedy strategy which is mentioned above, combined with techniques for quantifying changes in rewards, and weighted exploration. It also involves changing exploration rate, $\epsilon$, as well as learning rate, $\alpha$, dynamically as per the changes in rewards. 

In typical $\epsilon$-greedy approach, the random arm is selected with a small probability $\epsilon$, otherwise, the current optimal arm is selected. Since the optimal arm is selected, overall, at more time steps, we can say that the changes that are detected in rewards of optimal arm reflect more up-to-date fluctuations in the market demand in comparison to other arms. Therefore, we only focus on the current optimal arm for detecting changes. As agents can see only their own rewards, our change quantification mechanisms are based on changes in rewards. Unlike other works based on change detection \citep{liu2018change,ghatak2020change,cao2019nearly}, our algorithm quantifies the change, instead of just getting the signal when change is detected. Quantifying the changes help in making suitable modifications to learning rate, exploration rate, and weighted exploration. We quantify changes for two separate purposes; one for modifying $\alpha$ and $\epsilon$, and other for weighted exploration. For our calculations, we store values that are based on rewards in memory of length $M$; those are Q-values and the running averages of rewards. Let $\bar{\mu}$ be the mean of any of these two sets of values. We quantify the change with formula: $|(r^{k}_{t}-\bar{\mu})/\bar{\mu}|$, where $(r^{k}_{t}$ is the reward obtained by selecting arm $k$ (current optimal) at time $t$. The resulting value can be directly used as both $\alpha$, as well as $\epsilon$, given that it is within their pre-specified limits. We did not find much difference in results over which set of values we use. In the algorithm given below, we use Q-values in our calculations.

For weighted exploration, we take advantage of the specific nature of this problem, i.e., the ordered action-space, and that the significance of an action is dependent on its distance from the optimal action. While assigning weights to the actions, we use normal pdf (probability density function). For that we consider the normal distribution with mean at optimal action and standard deviation as a quantified change. Here, we quantify changes mainly in two ways: $|(r^{k}_{t}-(Q^{k}_{t})|$ and std. in Q-values (or the running averages) stored in memory for arm $k$ (used in algorithm below). In our observation, results are similar for using these techniques. Deriving std. in this way ensures that the focus range of exploration narrows down or spans out based on the degree of change. These techniques play crucial role in deriving agility of algorithm for changing the course of actions with varying demand. As the received reward deviates from expected (mean) reward, the exploration rate goes up and weights associated with the arms tend towards uniformity, and vice versa. This helps in promptly finding new optimal arm and sticking to it until changes in rewards are detected again.

\begin{algorithm}
\caption{Adaptive with Weighted Exploration (AWE) $\epsilon$-Greedy}\label{alg:algo}
\begin{algorithmic}

\State \textbf{Input:}
\State \hspace{5mm} $K>2$, number of arms
\State \hspace{5mm} $M\in Z$, memory length
\State \hspace{5mm} $\epsilon, \epsilon_{min}, \epsilon_{max} \in (0,1)$, exploration rate and its thresholds
\State \hspace{5mm} $\alpha, \alpha_{min}, \alpha_{max} \in (0,1)$, learning rate and its thresholds

\State Initialize $Q(k)$ to arbitrary values between $(0,1)$, for $k=1 \dots K$.
\State Initialize $flag \, = \, False$ (for tracking if current optimal action is taken)
\State Initialize $w(k)=Q(k)$, for $k=1 \dots K$.
 \For{$t\in\{1,\dots,T\}$}
 
    
    \State Sample $rNum \sim \mathcal{U}(0,\,1)$
    
     \If{$rNum \leq \epsilon$} 
    
        \State Play arm $k_{t}$ chosen from arms $1 \dots K$ according to weights $w_{t}(k)$ 
    
     \Else
    
        \State Play arm $k_{t} = argmax_{k}(Q_{t}(k))$
        \State $flag \, = \, True$
    \EndIf
    
    \State Receive reward $r^{k}_{t}$ calculated using eq. \ref{eq2}.
    \State Update $Q_{t}(k) = \alpha r^{k}_{t} + (1-\alpha) Q_{t-1}(k)$
    
     \If{$flag \, = \, True$} 
        \State Calculate \[ \bar{\mu} = \frac{1}{M}\sum_{i=t-M}^{t} Q_{i}(k)\]
    
        \State Calculate  \[ \hat{\mu} = argmax_{k}(Q_{t}(k))\]
        \State Calculate \[\hat{\sigma}=\frac{1}{M-1}\sum_{i=t-M}^{t} Q_{i}(k) -\bar{\mu} \]
        
         \For{$k\in\{1,\dots,K\}$}
         
            \State Update $w_{t+1}(k)=f(k|\hat{\mu},\hat{\sigma})$, where $f$ is normal probability density function
        \EndFor

        \State Calculate $newRate = |(r^{k}_{t}-\bar{\mu})/\bar{\mu}|$
        
         \If{$newRate \, < \epsilon_{min}$}
        
            \State $\epsilon \, = \, \epsilon_{min}$
        
         \ElsIf{$newRate \, > \, \epsilon_{max}$}
        
            \State $\epsilon \, = \, \epsilon_{max}$
        
         \Else
        
            \State $\epsilon \, = \, newRate$
        \EndIf
        
         \If{$newRate \, < \alpha_{min}$}
        
            \State $\alpha \, = \, \alpha_{min}$
        
         \ElsIf{$newRate \, > \, \alpha_{max}$}
        
            \State $\alpha \, = \, \alpha_{max}$
        
         \Else
        
            \State $\alpha \, = \, newRate$
        \EndIf

    \State $flag \, = \, False$
    \EndIf
    \EndFor

\end{algorithmic}
\end{algorithm}

\section{Experiments and Results}

We empirically evaluate our proposed algorithm AWE $\epsilon$-Greedy. We start with  simulations similar to the ones used in \citep{waltman2008q,xu2020reinforcement}. However, unlike these works, we assume non-stationary demand. Specifications such as number of agents, number of actions, values of parameters $v$ and $c$, as well as starting value for $u$ are inspired from those works. We further explore the scalability of our method by running simulations with models consisting of up to $50$ firms and up to $500$ actions. Here, a simulation means the entire process of a repeated Cournot game that lasts for several time steps. For each simulation, we compute the graph of average values calculated over $100$ time steps. We investigate the efficiency of our proposed approach in dealing with non-stationarity. We study individual as well as joint outputs produced by agents along the entire simulation. We compare these outcomes with three types of equilibria that are possible in Cournot models, i.e., collusive, Nash, and Walrasian equilibrium. We also analyze changes in outcomes caused by factors such as total agents present in the system, size of action space, and asymmetry in firms. We use adaptive $\epsilon$-greedy algorithm \citep{dos2017adaptive} as a baseline to compare our method. Similar to our approach, this algorithm is also based on $\epsilon$-greedy strategy and quantifies the changes in rewards to modify the exploration rate.

The notion of \emph{regret} is important in analyzing the results of bandit problems. In this paper, we consider a notion of regret that is specific to Cournot models. This notion is based on the assumption that the firms always strive for the maximum profit which can only be achieved if they form a cartel, i.e., full collusion. With this notion, regret can be seen as the difference between actual outcomes with those that are possible in collusive equilibrium. 



\subsection{Parameters}

We run each simulation for $100,000$ time steps. Here, simulation means the entire process of a repeated Cournot game that lasts for several time steps. We use the memory length of $10$ for all types of simulations. We found that results are moderately sensitive towards memory length, i.e., small changes do not affect the outcomes, but big changes can do so. However, we do not fine-tune it for every type of simulation. Modifications in learning rate $\alpha$ and exploration rate $\epsilon$ are confined to certain thresholds. The limits are defined as: $\alpha_{max} = 0.3$, $\alpha_{min} = 0.01$, $\epsilon_{max} = 0.3$, and $\epsilon_{min} = 0.05$. Similar to the memory length, these values are used for all the simulations, and outcomes are moderately sensitive to these hyper-parameters.

\subsection{Simulations with Cournot Duopoly Model}

\begin{figure*}
     \centering
     \begin{subfigure}[b]{0.33\textwidth}
         \centering
         \includegraphics[width=\textwidth]{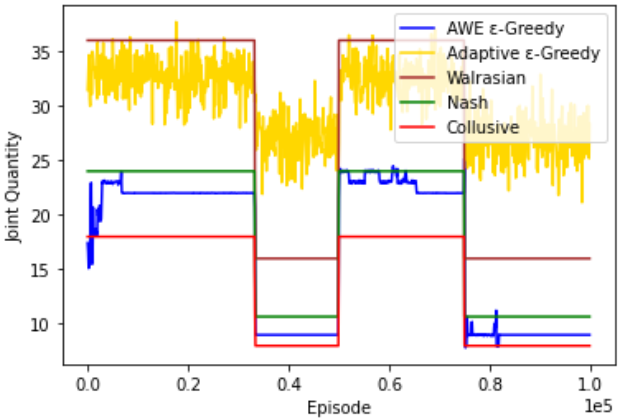}
         \caption{}
         \label{fig:adaptiveQ1}
     \end{subfigure}
     \hfill
     \begin{subfigure}[b]{0.33\textwidth}
         \centering
         \includegraphics[width=\textwidth]{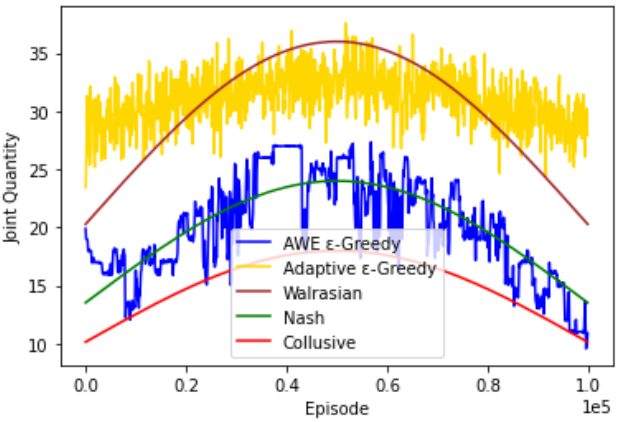}
         \caption{}
         \label{fig:adaptiveQ2}
     \end{subfigure}
     \hfill
     \begin{subfigure}[b]{0.33\textwidth}
         \centering
         \includegraphics[width=\textwidth]{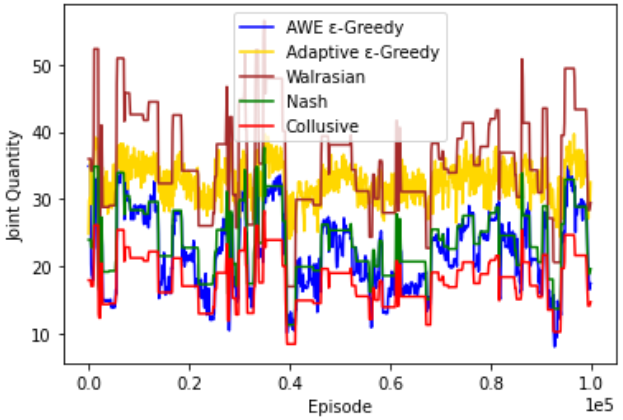}
         \caption{}
         \label{fig:adaptiveQ3}
     \end{subfigure}
        \caption{For Cournot duopoly model with $2$ firms: comparison of joint quantity obtained by AWE $\epsilon$-greedy and adaptive $\epsilon$-greedy algorithm, along with collusive, Nash and Walrasian equilibrium. Cournot model has symmetric firms with $n=2$, $K=40$, $c=4$, $u_{s}=40$, and $v=1$. Fig. \ref{fig:adaptiveQ1}, \ref{fig:adaptiveQ2}, and \ref{fig:adaptiveQ3} show results for demand pattern 1, 2, and 3, resp.}
        \label{fig:adaptiveQ}
\end{figure*}

\begin{figure*}
     \centering
     \begin{subfigure}[b]{0.33\textwidth}
         \centering
         \includegraphics[width=\textwidth]{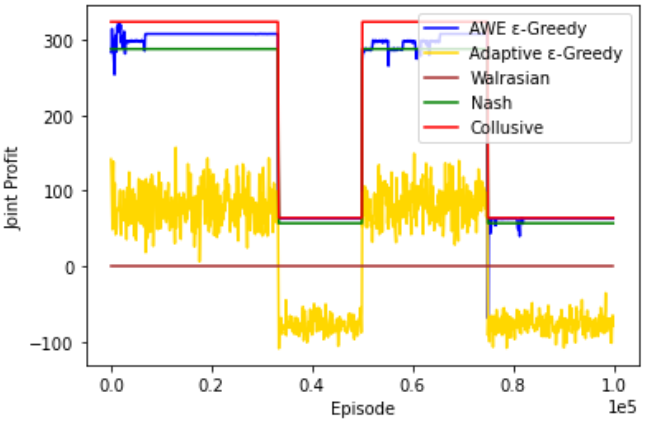}
         \caption{}
         \label{fig:adaptiveP1}
     \end{subfigure}
     \hfill
     \begin{subfigure}[b]{0.33\textwidth}
         \centering
         \includegraphics[width=\textwidth]{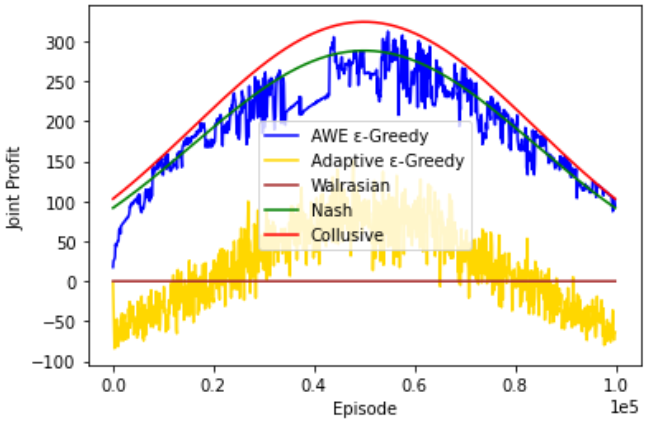}
         \caption{}
         \label{fig:adaptiveP2}
     \end{subfigure}
     \hfill
     \begin{subfigure}[b]{0.33\textwidth}
         \centering
         \includegraphics[width=\textwidth]{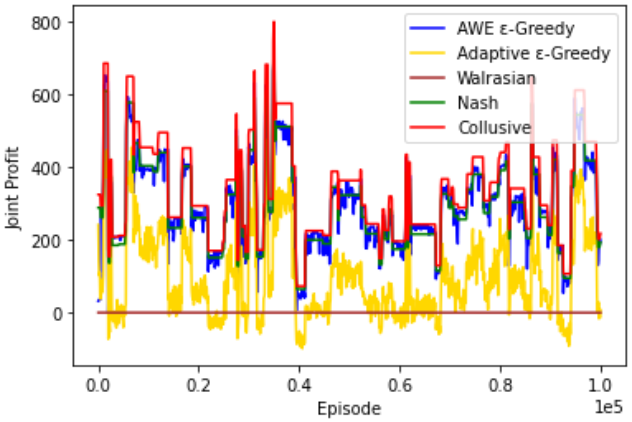}
         \caption{}
         \label{fig:adaptiveP3}
     \end{subfigure}
        \caption{For Cournot duopoly model with $2$ firms: comparison of joint profit obtained by AWE $\epsilon$-greedy and adaptive $\epsilon$-greedy algorithm, along with collusive, Nash and Walrasian equilibrium. Cournot model has symmetric firms with $n=2$, $K=40$, $c=4$, $u_{s}=40$, and $v=1$. Fig. \ref{fig:adaptiveP1}, \ref{fig:adaptiveP2}, and \ref{fig:adaptiveP3} show results for demand pattern 1, 2, and 3, resp.}
        \label{fig:adaptiveP}
\end{figure*}

We first consider simplistic Cournot model with $2$ firms. This model is inspired by simulations used in \citep{waltman2008q,xu2020reinforcement}. Similar to these works, we run simulations with the number of symmetric firms in the system varying from $2$ to $6$. However, we only present results of duopoly model in figs. \ref{fig:adaptiveQ} and \ref{fig:adaptiveP}. They used $u=40$, $v=1$ as constants in the inverse demand function (eq. \ref{eq1}), along with constant marginal cost, $c=4$. The action space is discrete, where agents can choose a production quantity between $0-40$. We compare our results with those obtained by using adaptive $\epsilon$-greedy. From figs. \ref{fig:adaptiveQ} and \ref{fig:adaptiveP}, we can see that the adaptive $\epsilon$-greedy perform much worse than our approach, especially in terms of adjusting to new market demand, even for simple duopoly model; this is the main reason we included results of Cournot duopoly model. Consequently, we do not include the comparison with adaptive $\epsilon$-greedy in other graphs.

Fig. \ref{fig:adaptiveQ} shows the comparison of joint quantities; it shows collective changes in the action course of agents. With adaptive $\epsilon$-greedy, agents are able to somewhat change the action course according to changes in demand, but the degrees/pattern of those changes does not match the demand pattern properly as compared to the nearly perfect alignment of outputs obtained by using our approach with the demand pattern. Moreover, with adaptive $\epsilon$-greedy, the overall outputs are closer to Walrasian equilibrium, and in some cases more than Walrasian equilibrium. This also reflects in fig. \ref{fig:adaptiveP}, where the joint profits for adaptive $\epsilon$-greedy go to negative values for some part of the simulations. With our approach, the outcomes are either collusive or either in Nash equilibrium, with some exceptions where those are between Nash and Walrasian equilibrium. By the notion of regret discussed above, we can see that our approach causes less regret than adaptive $\epsilon$-greedy as results obtained by it are farther from collusive equilibrium than the results obtained by our approach. We have also observed that the outputs produced by individual but identical firms in a symmetric Cournot model are mostly similar when the firms use our approach. It means that the results are fair to each agent, and individual agents can rely on our method to get a fair share of the market profit despite being autonomous and independent.

For all types of simulations, we can see that the sudden changes in demand sometimes cause a sharp drop in joint outputs, however, quick recovery towards matching new demand can also be seen. The scale of simulation seems to affect the degree of sharp drops and recovery from those. This is especially true for results with demand patterns $1$ and $3$ which consists of sudden shocks in demand. Interestingly, the type of non-stationarity does not seem to affect the resulting equilibrium in outcomes obtained by our approach.

\subsection{Scaling Number of Firms}

\begin{figure*}
     \centering
     \begin{subfigure}[b]{0.33\textwidth}
         \centering
         \includegraphics[width=\textwidth]{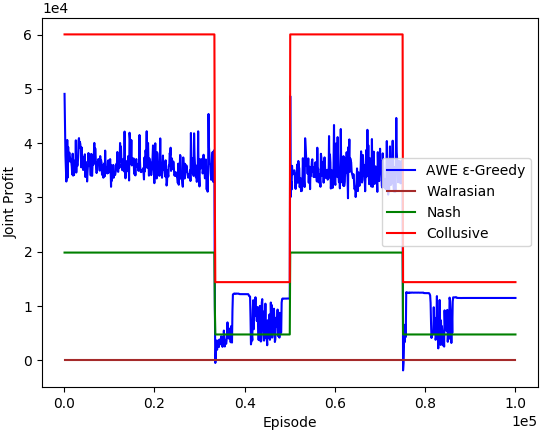}
         \caption{}
         \label{fig:sag10_1}
     \end{subfigure}
     \hfill
     \begin{subfigure}[b]{0.33\textwidth}
         \centering
         \includegraphics[width=\textwidth]{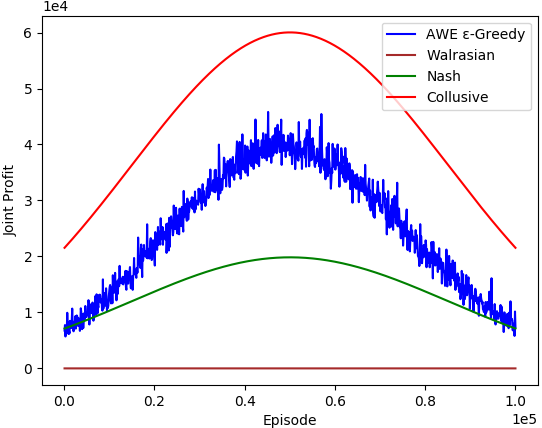}
         \caption{}
         \label{fig:sag10_2}
     \end{subfigure}
     \hfill
     \begin{subfigure}[b]{0.33\textwidth}
         \centering
         \includegraphics[width=\textwidth]{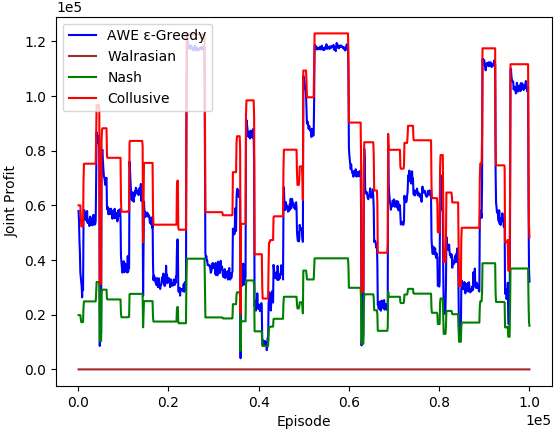}
         \caption{}
         \label{fig:sag10_3}
     \end{subfigure}
        \caption{For Cournot model with $10$ firms: comparison of joint profit obtained by AWE $\epsilon$-greedy with collusive, Nash and Walrasian equilibrium. Cournot model has symmetric firms with $n=10$, $K=50$, $c=10$, $u_{s}=500$, and $v=1$. Fig. \ref{fig:sag10_1}, \ref{fig:sag10_2}, and \ref{fig:sag10_3} show results for demand patterns 1, 2, and 3, resp.}
        \label{fig:sag10}
\end{figure*}

\begin{figure*}
     \centering
     \begin{subfigure}[b]{0.33\textwidth}
         \centering
         \includegraphics[width=\textwidth]{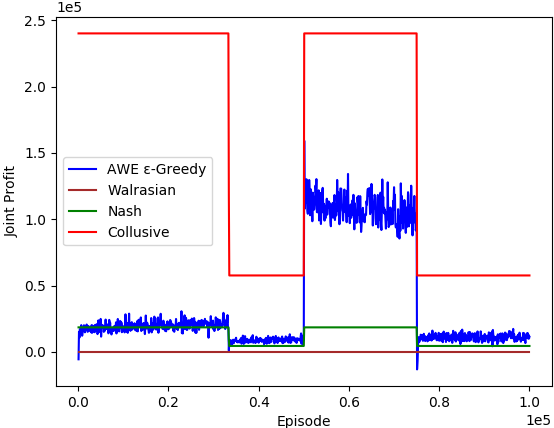}
         \caption{}
         \label{fig:sag50_1}
     \end{subfigure}
     \hfill
     \begin{subfigure}[b]{0.33\textwidth}
         \centering
         \includegraphics[width=\textwidth]{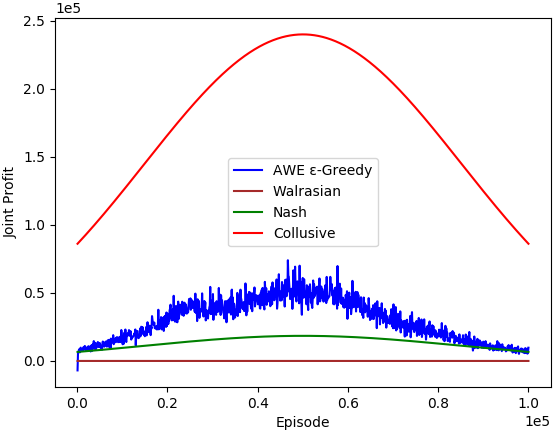}
         \caption{}
         \label{fig:sag50_2}
     \end{subfigure}
     \hfill
     \begin{subfigure}[b]{0.33\textwidth}
         \centering
         \includegraphics[width=\textwidth]{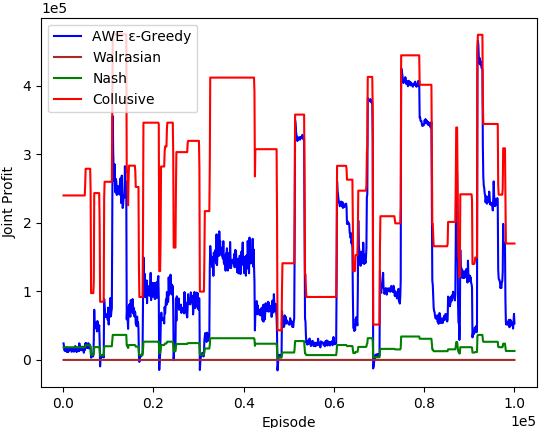}
         \caption{}
         \label{fig:sag50_3}
     \end{subfigure}
        \caption{For Cournot model with $50$ firms: comparison of joint profit obtained by AWE $\epsilon$-greedy with collusive, Nash and Walrasian equilibrium. Cournot model has symmetric firms with $n=50$, $K=50$, $c=20$, $u_{s}=1000$, and $v=1$. Fig. \ref{fig:sag50_1}, \ref{fig:sag50_2}, and \ref{fig:sag50_3} shows results for demand patterns 1, 2, and 3, resp.}
        \label{fig:sag50}
\end{figure*}

We check the scalability of our approach in terms of the total number of firms in a Cournot model. Cournot models with a large number of firms have been rarely explored in the literature. We use simulations with the varying number of firms, up to $50$ firms in a system. With the increasing number of firms, Nash equilibrium approaches Walrasian equilibrium. We show results for models with $10$ firms in fig. \ref{fig:sag10}, and with $50$ firms in fig. \ref{fig:sag50}. By comparing the two sets of results, we can see that with more number of firms in the system, there is less degree of collusion among them, and outcomes are either Nash or much closer to Nash. From fig. \ref{fig:sag50_1}, we can see that, for the same demand, our algorithm might perform bit differently in two separate time spans. Overall, these results suggest that with AWE $\epsilon$-greedy, firms can show collusive behavior even when there are large number of firms present in the market. This is in contrast with popular belief in the literature that with bigger markets, industry outputs tend to move towards Walrasian equilibrium.


\subsection{Scaling Action Space}
\begin{figure*}
     \centering
     \begin{subfigure}[b]{0.33\textwidth}
         \centering
         \includegraphics[width=\textwidth]{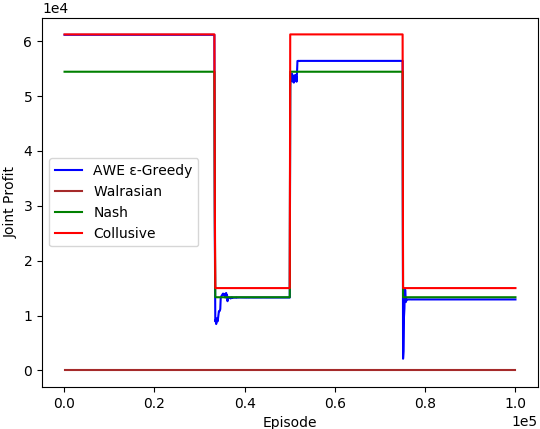}
         \caption{}
         \label{fig:sac1}
     \end{subfigure}
     \hfill
     \begin{subfigure}[b]{0.33\textwidth}
         \centering
         \includegraphics[width=\textwidth]{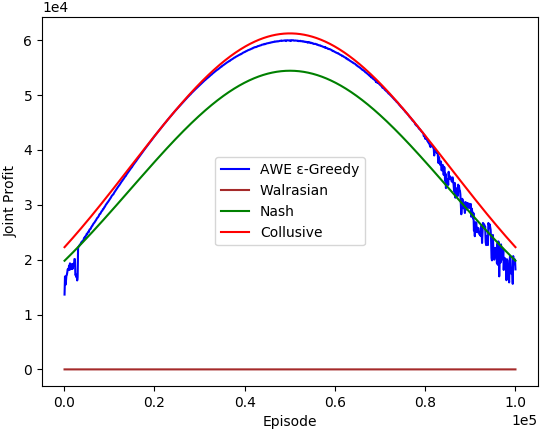}
         \caption{}
         \label{fig:sac2}
     \end{subfigure}
     \hfill
     \begin{subfigure}[b]{0.33\textwidth}
         \centering
         \includegraphics[width=\textwidth]{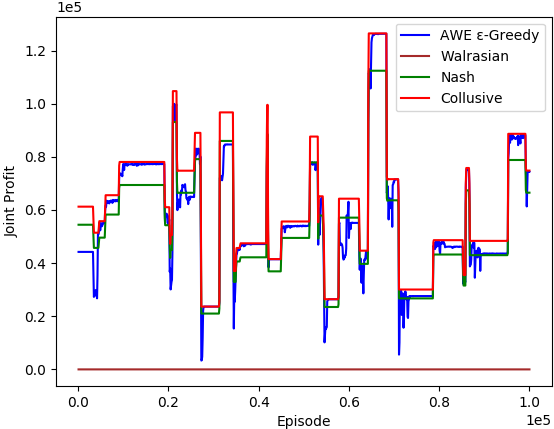}
         \caption{}
         \label{fig:sac3}
     \end{subfigure}
        \caption{For Cournot model with scaled actions: comparison of joint profit obtained by AWE $\epsilon$-greedy with collusive, Nash and Walrasian equilibrium. Corunot model has symmetric firms with $n=2$, $K=500$, $c=4$, $u_{s}=500$, and $v=1$. Fig. \ref{fig:sac1}, \ref{fig:sac2}, and \ref{fig:sac3} shows results for demand patterns 1, 2, and 3, resp.}
        \label{fig:sac}
\end{figure*}

We further check the scalability in terms of large action spaces as firms with high production capacities can exist in real-world markets. In fig. \ref{fig:sac}, we show the results for the Cournot duopoly model consisting of $2$ symmetric firms having action space size of $500$. Here, the baseline demand $u_{s}$ is also set to $500$. This is to promote the meaningful exploration of action space. From fig. \ref{fig:sac}, we can see that the degree of collusion is not affected by the increase in action space. However, especially from fig. \ref{fig:sac1} and \ref{fig:sac3}, it can be seen that the joint profit drops sharply towards Walrasian outcome when there are sudden changes in the demand, however, it quickly converges back to collusive or Nash outcomes. These drops are bigger than the ones observed in most of the other simulations; we think that increased exploration due to the large action space might be the reason behind it.

\subsection{Models with Asymmetric Firms}
\begin{figure*}
     \centering
     \begin{subfigure}[b]{0.33\textwidth}
         \centering
         \includegraphics[width=\textwidth]{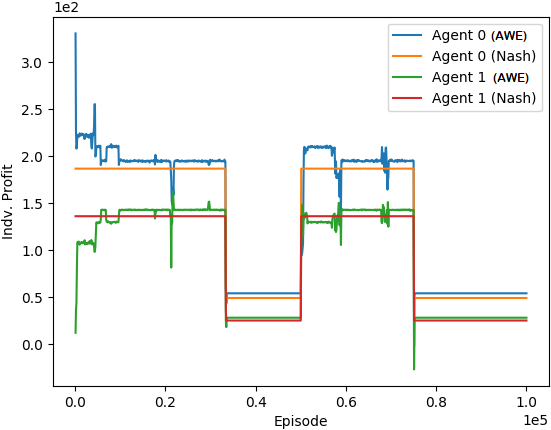}
         \caption{}
         \label{fig:ijp1}
     \end{subfigure}
     \hfill
     \begin{subfigure}[b]{0.33\textwidth}
         \centering
         \includegraphics[width=\textwidth]{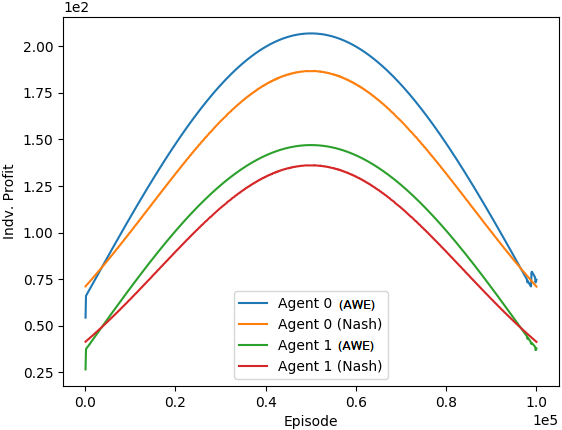}
         \caption{}
         \label{fig:ijp2}
     \end{subfigure}
     \hfill
     \begin{subfigure}[b]{0.33\textwidth}
         \centering
         \includegraphics[width=\textwidth]{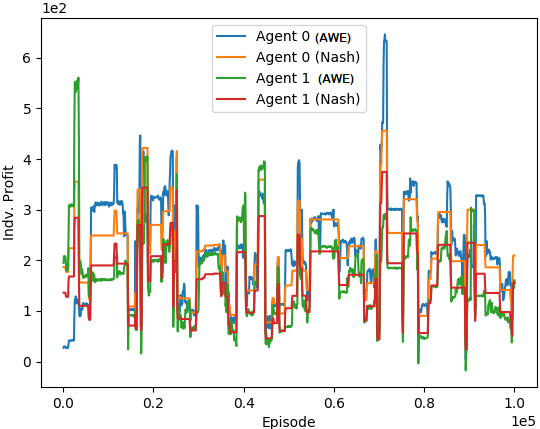}
         \caption{}
         \label{fig:ijp3}
     \end{subfigure}
        \caption{For Cournot duopoly model with $2$ asymmetric firms: comparison of  quantity obtained by individual agent using AWE $\epsilon$-greedy algorithm with respective Nash outputs. Corunot model has asymmetric firms with $n=2$, $K=40$, $c_{0}=1$, $c_{1}=3$, $u_{s}=40$, and $v=1$. Fig. \ref{fig:ijp1}, \ref{fig:ijp2}, and \ref{fig:ijp3} show results for demand patterns 1, 2, and 3, resp. (AWE) in graphs refer to AWE $\epsilon$-Greedy.}
        \label{fig:ijp}
\end{figure*}

We also study Cournot models with asymmetric firms, i.e., firms with different marginal costs. In simulations, we use Cournot duopoly models with two firms having different marginal costs. We show the results for Cournot duopoly model where the constant marginal cost of one firm is $1$, while that of another is $3$. We show comparison with Nash outcomes in fig. \ref{fig:ijp}. We can see that both the firms perform better than their respective Nash outcomes, i.e., showing collusive behavior. However, during further investigation, we found that as the difference between marginal costs increases or as the number of agents in the market increases, firms tend to produce highly diverse outcomes in terms of comparison with their respective Nash quantities. This may result in overall results being unfair towards some firms. Also, not all the firms are able to produce collusive or Nash outcomes.


\section{Conclusion}

We modeled the decision making in repeated Cournot games with non-stationary demand as the non-stationary bandit problem. Each firm separately faces its own instance of non-stationary MABs. We use three different non-stationary demand patterns to represent various kinds of non-stationarities found in real-world settings. We proposed AWE $\epsilon$-greedy algorithm, which incorporates mechanisms for quantifying changes in rewards, to help it adopt to varying demand. Outcomes obtained by our proposed approach show that it can help firms to readily and smoothly change their action course according to changing demand. Most of the outcomes are collusive, although full collusion is rarely seen. Scale of the model can affect the outputs. We have investigated scalability in terms of the total number of firms in the market, as well as the size of action spaces. We experimented with both symmetric and asymmetric types of firms in the market, and found that asymmetry in firms can cause unfair outcomes. Finally, we think that as we observed collusive behavior in many outcomes, the results are supportive of the concern over online algorithms being collusive without any external intervention \citep{timo2021ethics}.

\bibliographystyle{unsrtnat}
\bibliography{references}

\end{document}